# A More Accurate Approximation of Activation Function with Few Spikes Neurons

**Dayena Jeong**[1,2,*], **Jaewoo Park**[1,2,*], **Jeonghee Jo**[2], **Jongkil Park**[2], **Jaewook Kim**[2], **Hyun Jae Jang**[2], **Suyoun Lee**[2] and **Seongsik Park**[2,†]

[1] Seoul National University of Science and Technology, Seoul, Korea
[2] Korea Institute of Science and Technology, Seoul, Korea

enju0817@gmail.com, {t24264, jh.jo2, jongkil, jaewookk, hjjang, slee_eels, seong.sik.park}@kist.re.kr

**Objective:** Recent deep neural networks (DNNs), such as diffusion models [1], have faced high computational demands. Thus, spiking neural networks (SNNs) have attracted lots of attention as energy-efficient neural networks. However, conventional spiking neurons, such as leaky integrate-and-fire neurons, cannot accurately represent complex non-linear activation functions, such as Swish [2]. To approximate activation functions with spiking neurons, few spikes (FS) neurons were proposed [3], but the approximation performance was limited due to the lack of training methods considering the neurons. Thus, we propose tendency-based parameter initialization (TBPI) to enhance the approximation of activation function with FS neurons, exploiting temporal dependencies initializing the training parameters.

**Method:** Our method considers the temporal dependencies of parameters in FS neurons when setting initial values for training. Motivating with the fact that the spiking neurons should operate continuously, the parameters were initialized to have a temporal dependency. Our method mainly consists of three steps. The first step is pre-training, where local optimized parameters are obtained by training from random initial values. The parameters ($h(t)$, $d(t)$, $T(t)$ (Fig. 1)) represent the reset value subtracted from the membrane potential, the weight of output spikes, and the threshold, respectively. The second step is function fitting to obtain the temporal relationship of the pre-trained parameters, as shown in Fig. 2. The third step is to extract initial values of parameters from the fitted function at each time step t. Then, the parameters are trained with the initialization.

**Results:** We validated the proposed method with approximation of Swish at the neuron level and diffusion model at the network level. To evaluate fairly, we compared the approximation performance of random initial values, initial values with Gaussian noise added after pre-training, and our method. We also evaluated the performance at the network level with diffusion model with the trained neurons by each method. According to our experimental results, TBPI demonstrates more accurate approximation of Swish activation at the neuron level (Tab. 1), which leads to improved performance of diffusion model (Tab. 2).

**Conclusion:** TBPI improves the generalization in training of FS neurons by the parameter initialization, showing potential in other non-linear activation functions such as GELU that is used in Transformer architectures. Therefore, it will pave the way to energy-efficient artificial intelligence by enabling various deep learning models to be implemented with deep SNNs.

**Keywords**: FS neuron, approximation, training, Swish, Diffusion Models.

[*] These authors contributed equally.
[†] Corresponding author

**Poster Presentation**: Yes

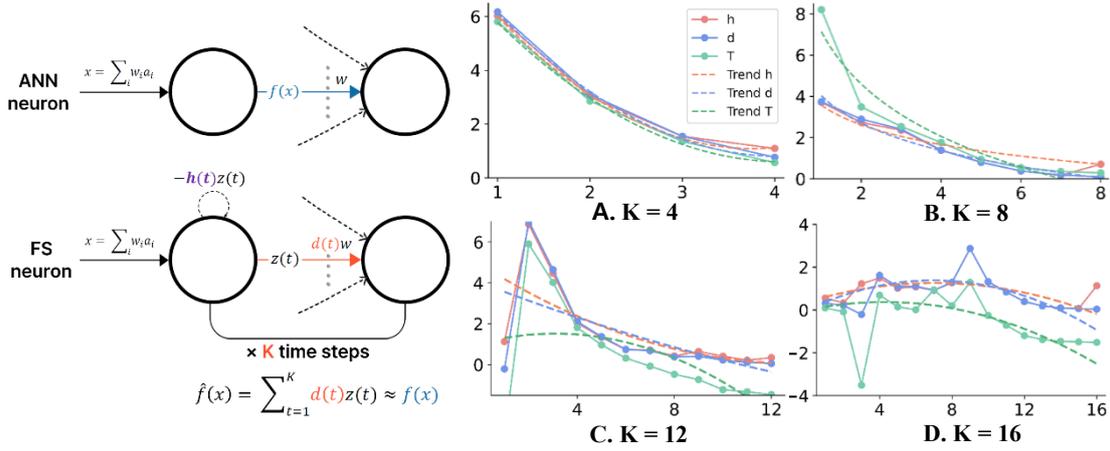

**Fig. 1.** An approximation $\hat{f}(x)$ of DNN's activation $f(x)$ with an FS neuron, where $x$, $w$ represent input and synaptic weight, respectively.

**Fig. 2.** The trained parameters of h(t), d(t), and T(t) depending on K (total time steps). Dashed lines indicate the fitted functions.

**Table 1.** MSE values for approximating the Swish function using three methods based on K are presented with a scale factor of $10^{-2}$.

| K<br>Time Steps | FS-Swish<br>Random Initial | FS-Swish<br>Gaussian Noise | TBPI-Swish<br>Temporal Noise |
|---|---|---|---|
| 4 | 3.8797 | 3.8796 | **3.8795** |
| 8 | 0.9776 | 0.8968 | **0.6295** |
| 12 | 0.1521 | 0.1517 | **0.1413** |
| 16 | 0.1425 | 0.1249 | **0.1246** |

**Table 2.** Training of the diffusion model was conducted on the Oxford Flower 102 dataset for 50 epochs in 4 separate experiments. The best performance from each experiment was recorded, and the average performance and standard error were subsequently calculated.

| Methods | K | Image Loss ↓ | KID ↓ | K | Image Loss ↓ | KID ↓ |
|---|---|---|---|---|---|---|
| ANN | - | 0.2263±0.0001 | 0.1542±0.0028 | | | |
| FS neuron | 4 | 0.2271±0.0003 | 0.1588±0.0012 | 12 | 0.2266±0.0002 | 0.1571±0.0007 |
|  | 8 | 0.2267±0.0001 | 0.1580±0.0020 | 16 | 0.2263±0.0001 | 0.1491±0.0021 |
| FS neuron<br>(Gaussian noise) | 4 | 0.2269±0.0003 | 0.1587±0.0040 | 12 | 0.2264±0.0001 | 0.1525±0.0021 |
|  | 8 | 0.2266±0.0001 | 0.1567±0.0022 | 16 | 0.2262±0.0001 | 0.1477±0.0010 |
| TBPI (ours) | 4 | 0.2279±0.0002 | 0.1584±0.0004 | 12 | 0.2261±0.0001 | 0.1500±0.0016 |
|  | 8 | 0.2264±0.0001 | 0.1545±0.0054 | 16 | **0.2257±0.0001** | **0.1474±0.0072** |

**References**

1. J. Song, C. Meng, S. Ermon.: Denoising Diffusion Implicit Models. In: International Conference on Learning Representations (2020)
2. Ramachandran, Prajit, Barret Zoph, and Quoc V. Le.: Searching for activation functions. arXiv preprint arXiv:1710.05941 (2017)
3. Stöckl, Christoph, and Wolfgang Maass.: Optimized spiking neurons can classify images with high accuracy through temporal coding with two spikes. Nature Machine Intelligence 3.3, 230-238 (2021)

Acknowledgements: This work was supported in part by the Korea Institute of Science and Technology (KIST) through 2E32960 and the National Research Foundation of Korea (NRF) grant funded by the Korea government (Ministry of Science and ICT) [NRF-2021R1C1C2010454].